\documentclass[lettersize,journal]{IEEEtran}
\usepackage{amsmath,amsfonts}
\usepackage{algorithmic}
\usepackage{algorithm}
\usepackage{array}
\usepackage[caption=false,font=normalsize,labelfont=sf,textfont=sf]{subfig}
\usepackage{textcomp}
\usepackage{stfloats}
\usepackage{url}
\usepackage{verbatim}
\usepackage{graphicx}
\usepackage{cite}
\usepackage[pagebackref=true,breaklinks=true,colorlinks,bookmarks=false]{hyperref}

\usepackage{dsfont}
\usepackage[cmyk]{xcolor}
\usepackage{multirow}

\newcommand{\etal}{\textit{et al}.}
\newcommand{\ie}[1]{{\textit{i.e.}{{#1}}}}
\newcommand{\eg}[1]{{\textit{e.g.}{{#1}}}}

\begin{document}

\title{Spatial Structure Constraints for Weakly Supervised Semantic Segmentation}

\author{Tao Chen, Yazhou Yao, Xingguo Huang, Zechao Li, Liqiang Nie and Jinhui Tang
	\thanks{Tao Chen, Yazhou Yao, Zechao Li and Jinhui Tang are with the School of Computer Science and Engineering, Nanjing University of Science and Technology, Nanjing 210094, China. (e-mail: taochen@njust.edu.cn; yazhou.yao@njust.edu.cn; zechao.li@njust.edu.cn; jinhuitang@njust.edu.cn).}
	\thanks{Xingguo Huang is with the College of Instrumentation and Electrical Engineering, Jilin University, Changchun 130061, China. (Email: xingguohuang@jlu.edu.cn).}
 \thanks{Liqiang Nie is with the School of Computer Science and Technology, Harbin Institute of Technology (Shenzhen), Shenzhen 518055, China. (Email: nieliqiang@gmail.com).}

	} 

\markboth{}%
{Shell \MakeLowercase{\textit{et al.}}: A Sample Article Using IEEEtran.cls for IEEE Journals}

\maketitle

\begin{abstract}
The image-level label has prevailed in weakly supervised semantic segmentation tasks due to its easy availability. Since image-level labels can only indicate the existence or absence of specific categories of objects, visualization-based techniques have been widely adopted to provide object location clues. Considering class activation maps (CAMs) can only locate the most discriminative part of objects, recent approaches usually adopt an expansion strategy to enlarge the activation area for more integral object localization. However, without proper constraints, the expanded activation will easily intrude into the background region. In this paper, we propose spatial structure constraints (SSC) for weakly supervised semantic segmentation to alleviate the unwanted object over-activation of attention expansion. Specifically, we propose a CAM-driven reconstruction module to directly reconstruct the input image from deep CAM features, which constrains the diffusion of last-layer object attention by preserving the coarse spatial structure of the image content. Moreover, we propose an activation self-modulation module to refine CAMs with finer spatial structure details by enhancing regional consistency. Without external saliency models to provide background clues, our approach achieves 72.7\% and 47.0\% mIoU on the PASCAL VOC 2012 and COCO datasets, respectively, demonstrating the superiority of our proposed approach. The source codes and models have been made available at \url{https://github.com/NUST-Machine-Intelligence-Laboratory/SSC}.
\end{abstract}

\begin{IEEEkeywords}
Semantic Segmentation, Weak Supervision, Image-Level Label, Spatial Structure Constraints.
\end{IEEEkeywords}

\section{Introduction}
\label{sec:intro}

With the development of deep learning algorithms, semantic segmentation has achieved remarkable progress and played an increasingly important role in many practical applications, such as autonomous driving and medical image analysis \cite{chen2023multi,ding2020semantic,he2021mgseg}. However, training a deep segmentation model requires a large number of high-cost pixel-wise accurate labels. Therefore, weakly supervised semantic segmentation (WSSS) has recently attracted the attention of researchers, which aims to alleviate the reliance on pixel-level annotations by resorting to weak supervision. Compared to bounding box \cite{dai2015boxsup,khoreva2017simple,jing2019coarse,song2019box}, point \cite{bearman2016s}, and scribble \cite{lin2016scribblesup,vernaza2017learning}, image-level label  \cite{kolesnikov2016seed,pei2022hierarchical,chen2021semantically,huang2018weakly,ahn2018learning,wei2018revisiting,jiang2019integral,chen2022saliency,li2022expansion,xu2023masked,huang2023paddles,huang2020low} is the most challenging annotation form which only indicates the presence or absence of certain classes of objects. It is also the most popular weak label due to its easy availability. In this paper, we concentrate on addressing the WSSS task based on image-level labels.

\begin{figure}[t]
	\begin{center}
		\includegraphics[width=\linewidth]{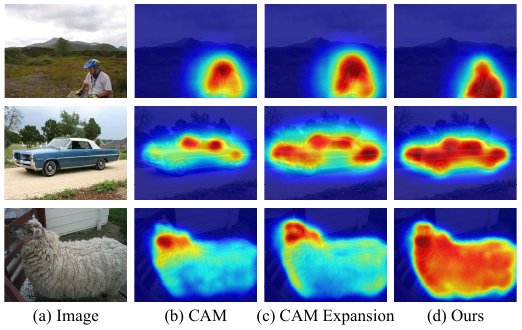}
	\end{center}
	\caption{Comparison between the traditional methods and ours. (a) Input image. (b) Localization maps produced by CAM \cite{zhou2016learning} only identify the most discriminative part of the object, \eg, the window of a car. (c) CAM expansion \cite{kim2021discriminative} results of traditional methods. They mainly focus on expanding the activation region and rely on saliency maps to provide background clues. Consequently, they will also inevitably result in over-activation, \ie, the expanded object activation intrudes into the background area. (d) Our results. Our proposed spatial structure constraints can constrain the activation within the object area to alleviate object over-activation (first and second rows) and help activate more integral object regions to mitigate object under-activation (last row). Best viewed in color.}
	\label{fig_moti}
\end{figure}

The typical pipeline of the image-level label based weakly supervised semantic segmentation is composed of three steps: 1) training a classification network with image-level annotations; 2) relying on visualization techniques to locate target objects and generating pixel-level pseudo-labels, \eg, class activation maps (CAMs) \cite{zhou2016learning}; 3) leveraging the obtained pseudo-labels to train a segmentation model in the fully-supervised setting. Though CAMs provide location clues for targets in images, they can only locate a small discriminative region of the target objects, which leads to unsatisfactory pseudo-labels with a high false negative rate. Therefore, recent efforts for WSSS tasks mainly focus on expanding the object activation to cover more integral object regions. For example, Kim \etal \cite{kim2021discriminative} proposed a discriminative region suppression approach to suppress the attention on discriminative regions and spread it to the adjacent non-discriminative area. They also leverage localization map refinement learning to further refine localization maps by recovering more missing parts.
Though these methods can enlarge activated object regions, they will also inevitably result in over-activation, \ie, the expanded object activation intrudes into the background area, illustrated in Fig.\,\ref{fig_moti}\,(c). Therefore, many approaches suffer from false positive activation and have to rely on additional saliency maps to provide background clues for post-processing or joint training. Few works focus on constraining the object attention during the expansion process.

In this paper, we propose to exploit spatial structure constraints (SSC) for weakly supervised semantic segmentation to alleviate the object over-activation problem mentioned above. Though CAMs can help locate regions of target objects, they can hardly provide any boundary clues due to the lack of shape information during training. Such amorphous localization accounts for the under-activation of naive CAMs and the over-activation of enlarged CAMs with attention expansion. Therefore, inspired by the self-supervised learning work of Autoencoder \cite{hinton2006reducing}, we propose a CAM-driven reconstruction module to directly reconstruct the input image from its CAM features \ie, the last-layer features used to generate CAMs. To endow CAMs with spatial structure information of the image content, we train the reconstruction network together with the classification backbone using a perceptual loss \cite{johnson2016perceptual} that depends on high-level features from a pre-trained loss network, instead of using the per-pixel loss function (L1 or L2 loss) depending only on low-level pixel information. It is noteworthy that we reconstruct images directly from the class-specific CAM features. This is essential to guide the classification network to generate CAMs that maintain the spatial structure of image content by penalizing the reconstructed image that semantically and spatially deviates from the corresponding input image.

However, as we reconstruct images from deep semantic features with reduced resolution, only overall image content and coarse spatial structure are preserved in CAMs but exact shapes are not. Therefore, we propose an activation self-modulation module to further refine CAMs with higher resolution spatial structure details via enhancing regional consistency. Specifically, we resort to superpixels that group pixels that are similar in color and other low-level features. We first refine the CAM features by averaging their values in each superpixel to obtain regional consistent features, which will also lead to CAMs with local smoothness and consistency. Then we align the CAMs to the obtained regional consistent CAM for encouraging the classification network to maintain regional consistency. However, the average operation will also significantly suppress the activation of the discriminative part if the high activation only occupies a small portion of the corresponding superpixel, which will lead to unwanted regional under-activation. Therefore, we also leverage a reliable activation selection strategy to maintain the high activation of the most discriminative region. Our proposed activation self-modulation module simultaneously enhances the regional consistency of CAM and keeps the reliable high activation, thus can significantly preserve the spatial structure details of the target objects. As can be seen in the first two rows of Fig.\,\ref{fig_moti}\,(d), our proposed spatial structure constraints can constrain the activation within the object area to alleviate the object over-activation problem mentioned above. Besides, as shown in the last row of Fig.\,\ref{fig_moti}\,(d), constraining the object attention with image content can also help activate more integral object regions to mitigate object under-activation not tackled by previous approaches.

Our proposed SSC is trained jointly with the classification network in a single round, which can be directly plugged into existing networks, as shown in Fig.~\ref{fig_framework}. Without external saliency models to provide background clues, our approach achieves 72.7\% and 47.0\% mIoU on the PASCAL VOC 2012 and COCO datasets, respectively. These results demonstrate the advantage of alleviating the object over-activation problem by exploiting spatial structure constraints. Our contributions can be summarized as follows:

\begin{itemize}
	
	\item We propose spatial structure constraints (SSC) for weakly supervised semantic segmentation to alleviate the object over-activation problem of CAM expansion. 
	
	\item To preserve coarse spatial structure of the image content, we propose a CAM-driven reconstruction module with the perceptual loss that directly reconstructs the input image from its CAM features.
	
	\item  We propose an activation self-modulation module with a reliable activation selection strategy to further refine CAMs with finer spatial structure details by enhancing regional consistency.

\end{itemize}

The rest of this paper is organized as follows: 
the related work is described in Section \ref{related_work} and our approach is introduced in Section \ref{approach}; we then report our evaluations and ablation studies on two widely-used datasets in Section \ref{experiments} and finally conclude our work in Section \ref{conclusion}.

\section{Related Work}
\label{related_work}

\subsection{Weakly Supervised Semantic Segmentation}
Semantic segmentation aims to identify the category of each image pixel. It has achieved significant progress with the development of deep learning. For example, while prevalent works like DeepLabV2 \cite{chen2017deeplab} typically learn parametric class representations, the work of \cite{zhou2022rethinking} represents each class as a set of non-learnable prototypes to achieve dense prediction by nonparametric nearest prototype retrieving. Recently, LogicSeg \cite{li2023logicseg} integrates neural inductive learning and logic reasoning into a structured parser to holistically interpret visual semantics. However, collecting dense annotations for training fully supervised segmentation models is labor-intensive and time-consuming.

Compared to pixel-level accurate labels, weakly supervised semantic segmentation resorts to weaker annotations for alleviating the annotation burden of the segmentation task, \eg, image-level label  \cite{kolesnikov2016seed,pei2022hierarchical,chen2021semantically,huang2018weakly,ahn2018learning,wei2018revisiting,jiang2019integral,chen2022saliency,li2022expansion,xu2023masked,huang2023paddles,huang2020low}, bounding box \cite{dai2015boxsup,khoreva2017simple,song2019box}, point \cite{bearman2016s}, and scribble \cite{lin2016scribblesup,vernaza2017learning}. Among them, the image-level label is the most popular one and has attracted much attention of researchers. Visualization techniques like CAMs \cite{zhou2016learning} are widely adopted to leverage image-level tags to locate target object regions for facilitating the generation of pixel-level pseudo labels. Considering object activation in CAMs is usually incomplete and lacks shape information, recent works mainly focus on: (1) expanding object attention to cover more integral target regions, and (2) refining CAMs for accurate boundaries.

An effective way for attention expansion is to transfer the surrounding discriminative information to non-discriminative object regions. For example, dilated \cite{wei2018revisiting} and deformable \cite{li2022expansion} convolutions are exploited to sample increasingly less discriminative object regions and Kim \etal \cite{kim2021discriminative} propose a discriminative region suppression approach. Adversarial erasing is proposed to hide highly activated regions and drive the network to find more object parts \cite{wei2017object,tu2023learning}. Instead of treating each image independently, recent researchers also propose group-wise learning to mine cross-image semantic relations \cite{zhou2021group,wang2022looking}. While these methods can discover more object parts, they suffer from the object over-activation problem and rely on additional saliency maps to provide background clues \cite{jiang2019integral,kim2021discriminative,lee2021railroad,jiang2022l2g,zhao2021salience}. The generation of saliency maps usually involves other datasets to train a salient object detection model. In contrast, in this paper, we propose spatial structure constraints for weakly supervised semantic segmentation to constrain the activation of CAM expansion.

For CAM refinement, Ahn and Kwak propose AffinityNet \cite{ahn2018learning} to predict semantic affinity between a pair of adjacent image coordinates, which are then propagated via a random walk to recover delicate shapes of objects. To learn semantic affinities more effectively and efficiently, class boundary detection is exploited in IRN \cite{ahn2019weakly} to discover the entire instance areas with more accurate boundaries. The work of BES \cite{chen2020weakly} proposes explicitly exploring object boundaries through learning with synthetic boundary labels derived from localization maps. In this paper, we propose CAM-driven reconstruction and activation self-modulation modules, which can refine CAMs with more accurate boundaries by exploiting both coarse and fine-level spatial structure information of the input image.

\subsection{Self-Supervised Learning}
Deep supervised learning algorithms typically require large-scale annotated data to achieve satisfactory performance. In contrast, self-supervised learning (SSL), a branch of unsupervised learning methods, aims to learn general image and video features from large numbers of unlabeled examples \cite{jing2020self,chen2021enhanced,gui2023survey}. A pre-defined pretext task is usually designed for SSL with pseudo labels automatically generated based on attributes of data, \eg, relative position \cite{doersch2015unsupervised}, jigsaw \cite{noroozi2016unsupervised}, and rotation \cite{komodakis2018unsupervised}. Recently, through constructing the pretext task of instance discrimination, contrastive learning-based SSL methods, such as MoCo and SimCLR series \cite{he2020momentum,yao2021jo,yao2017exploiting,chen2020simple,chen2020big}, have pushed the performance of self-supervised pretraining to a level comparable to that of supervised learning. With the advent of vision transformer, masked image modeling (MIM) methods like MAEs \cite{he2022masked} and SimMIM \cite{xie2022simmim} have also become very popular due to their massive potential in SSL. Following the idea of Autoencoder \cite{hinton2006reducing},  the pioneering work about image generation-based SSL methods, we propose a CAM-driven reconstruction module to reconstruct the input image from features. However, different from the goal of Autoencoder to reduce the image dimension or extract features for downstream tasks, our reconstruction module aims to preserve the spatial structure of the image content. Therefore, we propose to directly reconstruct the input image from its CAM features and leverage perceptual loss to train the networks.

\subsection{Superpixel Guidance in Weakly Supervised Segmentation}
Superpixel aims to group pixels that are similar in color and other low-level features. After obtaining visually significant entities, superpixels can lower the number of primitives required for future processing stages and thus have been widely applied in tasks like medical and satellite imaging \cite{kumar2023extensive}. Weakly supervised approaches also explore the contexts of superpixels to improve the segmentation performance \cite{xing2016weakly,li2020weakly,fan2020learning,yi2022weakly}. For example, Li \etal \cite{li2020weakly} apply a superpixel-CRF refinement model to rectify the mistakes of the initial pixel annotations by an iteration framework. SGWS \cite{yi2022weakly} introduces the local and global consistency information of superpixel to optimize the coarse initial seeds that only cover the most discriminative parts of objects. Different from these methods that leverage superpixels at the post-processing stage to refine pseudo labels, we propose to exploit their regional consistency during the training of the classification network to derive more integral object activation. The most related work is ICD \cite{fan2020learning}, which leverages superpixels to refine the ICD score for intra-class discrimination. 
However, focusing on separating the foreground and the background pixels, ICD \cite{fan2020learning} is mainly learned by single-class images to avoid mixing multi-class objects, which hampers its application in complex datasets like COCO \cite{lin2014microsoft}. In contrast, our proposed spatial structure constraints are conducted directly with the CAM feature and thus can be extended to multi-class images. Besides, our proposed activation self-modulation keeps the reliable high activation rather than simply learning from the regional consistency.

\begin{figure*}[t]
	\begin{center}
		\includegraphics[width=\linewidth]{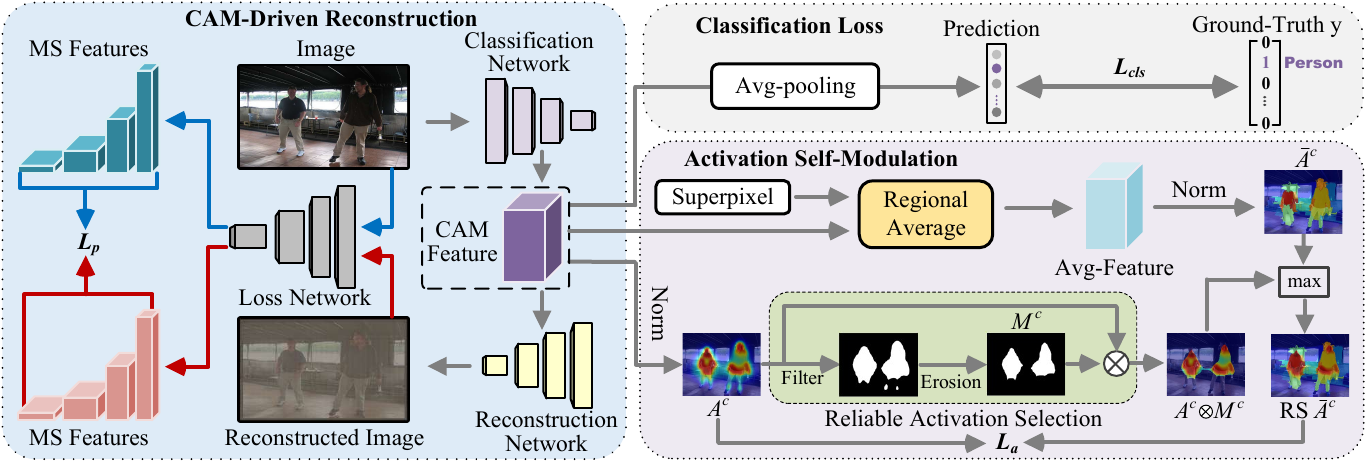}
	\end{center}
	\caption{The architecture of our proposed approach. While training the classification network with the image-level labels, we propose a CAM-driven reconstruction module to reconstruct the input image from its CAM-related features. Moreover, we propose an activation self-modulation module to further refine CAMs with finer spatial structure details through enhancing regional consistency. Our proposed modules help the classification network learn to preserve the spatial structure of the image content and constrain high activation within the object area. $\otimes$ is the Hadamard product. Best viewed in color.}
	\label{fig_framework}
\end{figure*}

\section{The Proposed Approach}
\label{approach}

In this paper, we propose spatial structure constraints (SSC) for weakly supervised semantic segmentation to alleviate the object over-activation problem induced by CAM expansion. Our framework is illustrated in Fig.~\ref{fig_framework}. Given only image-level weak labels, a classification network is trained. Then CAMs are derived from the last-layer convolutional features to help locate the object regions and generate pixel-level pseudo labels for standard segmentation network training. Considering existing attention expansion methods suffer from the over-activation problem, we first introduce a CAM-driven reconstruction network to reconstruct the input image from its CAM-related features. This helps the classification network learn to preserve the coarse spatial structure of the image content and constrain high activation within the object area. In addition, we propose an activation self-modulation module to further refine CAMs with finer spatial structure details by enhancing regional consistency.

\subsection{CAM Generation and Attention Expansion}
Following the recent progress of WSSS approaches, we train a classification network and leverage CAMs to locate the target objects. To facilitate the implementation of our proposed CAM-driven reconstruction and activation self-modulation modules, we remove the final fully-connected layer in the classification network and set the output channel of the last convolutional layer to the number of classes $C$. We can thus directly generate object localization maps from the class-aware feature maps $F$ of the last convolutional layer in the forward pass, which is proven by \cite{zhang2018adversarial} identical to the attention generation process in the original CAMs \cite{zhou2016learning}. In detail, for each target category $c$, we feed attention map $F^{c}$ into a ReLU layer and then normalize it to range from 0 to 1:
\begin{equation}
	A^{c}=\frac{ReLU\left ( F^{c} \right )}{\max\left ( F^{c} \right )}.
	\label{eq_norm}
\end{equation}

For the training of the classification network, we adopt the multi-label soft margin loss as follows:
\begin{equation}
	L_{cls}=-\frac{1}{C} \sum_{c=1}^{C} y^{c} \log \sigma\left(q^{c}\right)+\left(1-y^{c}\right) \log \left[1-\sigma\left(q^{c}\right)\right].
\end{equation}
Here, $\sigma\left(\cdot \right)$ is the sigmoid function. $y^{c}$ is the image-level label for the $c$-th class. Its value is 1 if the class is present in the image; otherwise, its value is 0.

Since the classification network tends to only identify patterns from the most discriminative parts for recognition, the generated object activation is usually sparse and incomplete. Therefore, we follow the recent work of DRS \cite{kim2021discriminative} to expand the activated object regions, which is a simple yet effective method and requires few or no additional parameters. Given an intermediate feature map $X\in \mathds{R}^{K \times H \times W}$, where $H$, $W$, and $K$ are the height, width, and number of channels of $X$, we first leverage global max pooling to extract the max-element of each channel $X_{max}\in \mathds{R}^{K \times 1 \times 1}$. Then we suppress the discriminative regions by using a constant value of 0.55 (suggested in DRS \cite{kim2021discriminative}) to determine the upper bound of $X$.
\begin{equation}
	X_{new}=\min(X, 0.55 \cdot X_{max}).
\end{equation}
Such a parameter-free module suppresses the attention on the most discriminative regions and forces the classification network to focus on more non-discriminative regions.

\subsection{CAM-Driven Reconstruction}
While suppressing the highest activation can successfully spread the attention on discriminative regions to the surrounding non-discriminative object regions, it will also cause the over-activation problem. Without proper constraints, attention will easily intrude background area and parts of objects belonging to other categories. Therefore, many recent approaches resort to additional saliency maps to provide background clues for post-processing or joint training. However, the usage of saliency maps usually requires extra training of the salient object detection model with corresponding ground truth. In contrast, we propose a CAM-driven reconstruction module to help constrain the activation within the target object area. Specifically, we propose to leverage a reconstruction network to recover input images $I$ directly from the class-specific CAM features $F$ of the last convolutional layers:
\begin{equation}
	\hat{I} = Reconstruct(F),
\end{equation}
where $\hat{I}$ is the reconstructed image. Note that all channels of the CAM features are used here to reconstruct the image (\eg, 20-channel features for the VOC dataset) rather than one or several specific feature channels related to the categories that exist in the image.
For driving the classification network to preserve the spatial structure of image content, we penalize the reconstructed image that semantically and spatially deviates from the original image. To encourage the network to concentrate on the semantic content and spatial structure more than the color or texture of images, we propose to train the reconstruction network together with the backbone using a perceptual loss. This loss relies on high-level features from a pre-trained loss network rather than the per-pixel loss function depending only on low-level pixel information. Specifically, after obtaining the reconstructed image $\hat{I}$, we input it together with the original input image $I$ into a pre-trained loss network $\phi$. Let $\phi_{j}(I)$ and $\phi_{j}(\hat{I})$ be the loss network's feature maps of $j$-th convolutional layer (stage) output with shape  $C_{j}\times H_{j} \times W_{j}$, the feature reconstruction loss is the Mean Absolute Error between the two feature representations:
\begin{equation}
	\ell_{j}^{\text {rec}}(\hat{I}, I)=\frac{1}{C_j H_j W_j}\left\|\phi_j(\hat{I})-\phi_j(I)\right\|.
\end{equation}
The perceptual loss is then defined as the sum of several reconstruction losses for multi-stage output features of the loss network:
\begin{equation}
	L_{p} =\sum_{j=1}^{J}\frac{1}{2^{(J+1)-j} } \cdot  \ell_{j}^{\text {rec}}(\hat{I}, I).
\end{equation}
Here $J$ denotes the number of stages for reconstruction loss calculation. $\frac{1}{2^{(J+1)-j} }$ is the weight (designed to match the gradually reduced feature resolution) that controls the relative importance of reconstruction loss for different-stage features. While the higher level features contain rich semantic information, the lower level ones can provide rich clues of object edge and shape. Benefiting from the preserved spatial structure of image content, our CAM-driven reconstruction module can help constrain the activation of target objects (within their own region) from intruding into their surrounding background area. At the same time, the proposed CAM-driven reconstruction also promotes the activation to cover more integral object regions under the situation of object under-activation.

\subsection{Activation Self-Modulation}

Benefiting from the reconstruction module with perceptual loss, the activation derived from the class-aware features tends to be more compact and consistent to align with the content of the image. However, as we reconstruct images from deep semantic features with reduced resolution, only overall image content and coarse spatial structure are preserved in CAMs but exact shapes are not. To endow CAM with more detailed spatial structure knowledge, we propose an activation self-modulation module to further refine CAMs via enhancing regional consistency. For regional consistent activation learning, we first refine the CAM features $F$ by averaging their values in each superpixel $S_{k}$ to obtain regional consistent feature representations:
\begin{equation}
	\bar{F}_{ij} =\frac{ {\sum_{m,n\in S_{k(ij)}}} F_{mn}}{\left |S_{k(ij)}  \right | }.
\end{equation}
Here, $S_{k(ij)}$ denotes the $k$-th superpixel that contains the pixel at position $(i,j)$. $\left | \cdot \right |$ is the operation that counts the number of pixels in $S_{k(ij)}$. Then we apply the ReLU and maximum normalization illustrated in Equation \ref{eq_norm} to generate the regional consistent CAM. After that, we propose to align the CAM $A^{c}$ to the obtained regional consistent CAM $\bar{A}^{c}$ with a mean-squared loss for driving the classification network to extract features maintaining the regional consistency:
\begin{equation}
	L_{a} =\frac{1}{C^{'}H^{'}W^{'}} \sum_{c=1}^{C^{'}} \left\|A^{c}-\bar{A}^{c}\right\|^2,
\end{equation}
where $H^{'}, W^{'}$, and $C^{'}$ are the height, width of the CAM, and the number of classes that exist in the image. Note that we up-sample the CAM here to match the half resolution of the superpixel for taking advantage of finer spatial structure information. With our regional consistent activation learning, we can increase the attention of non-discriminative object areas, especially those within the superpixel that contain the most discriminative object part.  

However, the average operation will also dilute the activation of the discriminative part if it only occupies a small portion of the corresponding superpixel, resulting in unwanted regional low-activation. In the scenario where several sparse high activation regions exist, the original high attention within a large superpixel might be significantly suppressed after the average dilation. This will lead to the false modulation towards the opposite direction, and the localization map might gradually lose the activation of the object region. Therefore, we propose a reliable activation selection strategy to maintain the high activation of the most discriminative regions. Specifically, we first leverage an object threshold $T_{obj}$ to filter out the area with relatively higher attention values. Then we apply an erosion operation to further narrow the region mask $M^{c}$ with high activation for reliable activation selection. Our erosion operation can help remove the unwanted high activation intruding into the background. Consequently, we obtain the CAM with reliable activation selection (RS $\bar{A}^{c}$) for alignment as follows:
\begin{equation}
	\bar{A}^{c} =\max(\bar{A}^{c}, A^{c}\otimes M^{c}),
\end{equation}
where $\otimes$ is the Hadamard product. Our proposed activation self-modulation module simultaneously enhances the regional consistency of CAM and keeps the reliable high activation, thus can significantly preserve the spatial structure details of target objects.

\subsection{Training Objective}
With our proposed CAM-driven reconstruction and activation self-modulation modules, the overall training loss of the classification network is as follows:
\begin{equation}
	L=L_{cls} +\beta _{p}L_{p} + \beta _{a}L_{a}.
	\label{eq_all}
\end{equation}
Here, $\beta _{p}$ and $\beta _{a}$ are the hyperparameters that control the relative importance of perceptual loss and alignment loss. After leveraging the trained classification network to obtain CAMs to locate target objects, we follow the recent works \cite{liu2020weakly,zhang2020causal,lee2021anti,lee2021reducing,chen2022class} and adopt IRN \cite{ahn2019weakly} to further refine CAMs for pseudo label generation.

\section{Experiments}
\label{experiments}

\subsection{Datasets and Evaluation Metrics}
Following previous works, we evaluate our approach on the PASCAL VOC 2012 dataset \cite{everingham2010pascal} and COCO dataset \cite{lin2014microsoft}. As the most popular benchmark for WSSS, the PASCAL VOC 2012 dataset contains 21 classes (20 object categories and the background) for semantic segmentation. The official dataset split contains 1,464 images for training, 1,449 for validation and 1,456 for testing. Following the common protocol in semantic segmentation, we expand the training set to 10,582 images with additional data from SBD \cite{hariharan2011semantic}. The COCO dataset is a more challenging benchmark with 80 semantic classes and the background. Following previous works \cite{wang2020weakly,Li2021GroupWiseSM,zhang2020causal}, we use the default train/val splits (80k images for training and 40k for validation) in the experiment. For all the experiments, we only adopt the image-level classification labels for training. Mean intersection over union (mIoU) is adopted as the metric to evaluate the quality of our generated CAMs, pseudo labels and segmentation results. The results for the PASCAL VOC test set are obtained from the official evaluation server.

\subsection{Implementation Details}
For the classification network, we follow the work of IRN \cite{ahn2019weakly} and adopt the ResNet-50 \cite{he2016deep} model as our backbone for a fair comparison, which is pre-trained on ImageNet \cite{deng2009imagenet}. A $1\times1$ convolutional layer of $C$ channels is adopted as the pixel-wise classifier to generate CAM features. Then, the CAM features are directly input into our CAM-driven reconstruction module, and coarse spatial structure constraint on features is added through minimizing the perceptual loss between the reconstructed and the original images. On the other hand, CAM features are also fed into our activation self-modulation module to further refine CAMs with finer spatial structure details by enhancing regional consistency with an alignment loss. The momentum and weight decay of the SGD \cite{bottou2010large} optimizer are 0.9 and $1 \times 10^{-4}$. The initial learning rate is set to $0.1$. The input images are resized and cropped to $512\times512$, and the size of generated CAM feature is $32\times32$ with stride=16. We train the classification network for 10 epochs with batch size = 6.

The reconstruction network comprises one head ConvBlock, one ResBlock, four Up-sample blocks and one tail ConvBlock to recover the size of CAM features to the input image. Specifically, each Up-sample block contains a 4$\times$4 transpose convolution to up-sample the features and a 3$\times$3 convolution to further aggregate the representation. Its detailed architecture is demonstrated in Table \ref{tab_arc}. For the loss network, we adopt the VGG-19 model \cite{simonyan2014very} pre-trained on ImageNet \cite{deng2009imagenet}. For the perceptual loss, we set $J$ = 5 to choose features before each pooling layer for reconstruction loss calculation. For the superpixels, we directly leverage the ones provided by the work of ICD \cite{fan2020learning} for fair comparison. Specifically, superpixels are first generated with the method of \cite{felzenszwalb2004efficient} and then hierarchically merged with selective search \cite{uijlings2013selective} so that each image contains at most 64 superpixels.  

\begin{table}[t]
	
	\setlength{\tabcolsep}{1mm}

	\renewcommand\arraystretch{1.1}
	\centering
	\caption{Detailed architecture of the reconstruction network.}
	\begin{tabular}{{c|c|c}}
		\hline
		Num  &Block &Specification  \\
		\hline
		1&ConvBlock&3$\times$3 Conv, LayerNorm, Relu\\
		\hline	
		\multirow{2}{*}{1}&\multirow{2}{*}{ResBlock}&3$\times$3 Conv, LayerNorm, Relu\\
		&&3$\times$3 Conv, LayerNorm\\
		\hline	
		\multirow{2}{*}{4}&\multirow{2}{*}{Up-sample}&4$\times$4 ConvTranspose, InstanceNorm, Relu\\
		&&3$\times$3 Conv, LayerNorm, Relu\\
		\hline	
		1&ConvBlock&3$\times$3 Conv, Tanh\\
		\hline	
	\end{tabular}
	
	\label{tab_arc}	
\end{table}

For the second stage training of WSSS, following the recent work of BECO \cite{rong2023boundary}, we adopt DeeplabV2 \cite{chen2017deeplab} as the segmentation network, which uses ResNet101 \cite{he2016deep} as the backbone with an output stride of 16. The momentum and weight decay of the SGD optimizer are 0.9 and $10^{-4}$. The initial learning rate is set to $10^{-2}$ and is decreased using polynomial decay. The segmentation model is trained for 80 epochs and 40 epochs on VOC and MS COCO datasets, respectively, with a common batch size of 16. We also follow the default setting of DeeplabV2 \cite{chen2017deeplab} and conduct experiments with VGG16 backbone for a more comprehensive comparison with previous approaches \cite{ahn2018learning,fan2020learning,chen2020weakly,sun2021ecs,lee2021railroad,jiang2022l2g}.  Both ResNet101 and VGG16 backbones are pretrained on ImageNet dataset \cite{deng2009imagenet}.

\begin{table}[t]
	
	\setlength{\tabcolsep}{4mm}
	\renewcommand\arraystretch{1.1}
	\centering
 	\caption{Accuracy (mIoU) of pseudo-masks evaluated on PASCAL VOC 2012 training set.}
	\begin{tabular}{{lcc}}
		\hline
		Methods  & seed & w/ IRN \cite{ahn2019weakly} \\
		\hline
		IRN \cite{ahn2019weakly}$_{\text{CVPR19}}$&48.3&66.3\\
		MBMNet\cite{liu2020weakly}$_{\text{ACMMM20}}$&50.2&66.8\\
		CONTA \cite{zhang2020causal}$_{\text{NIPS20}}$&48.8&67.9\\
		AdvCAM \cite{lee2021anti}$_{\text{CVPR21}}$&55.6&69.9\\
		RIB\cite{lee2021reducing}$_{\text{NIPS21}}$&56.6&70.6\\
		ReCAM \cite{chen2022class}$_{\text{CVPR22}}$&56.6&70.5\\
		ESOL\cite{li2022expansion}$_{\text{NIPS22}}$&53.6&68.7\\
		D2CAM\cite{wang2023treating}$_{\text{ICCV23}}$&58.0&71.4\\
		\hline
		\textbf{SSC (Ours)} &\textbf{58.3}& \textbf{71.9}\\
		\hline	
	\end{tabular}
	\label{tab_pseudo}	
\end{table}

\subsection{Comparisons to the State-of-the-arts}

\textbf{Accuracy of Pseudo-Masks.} For weakly supervised semantic segmentation, the quality of generated pseudo labels directly influences the performance of the trained segmentation network. Therefore, we first present the comparison of the quality of pseudo-masks derived from our approach and other state-of-the-arts. As shown in Table \ref{tab_pseudo}, the segmentation seed generated by our approach can arrive at the mIoU of 58.3\%, bringing a gain of 10\% compared to the baseline reported by IRN \cite{ahn2019weakly}. Our method can obtain more accurate seeds than state-of-the-art methods like RIB\cite{lee2021reducing} and ReCAM \cite{chen2022class} by 1.7\%. Our proposed SSC can also outperform the recent D2CAM\cite{wang2023treating} by 0.3\% mIoU. Leveraging the random walk algorithm in IRN \cite{ahn2019weakly} to further refine CAMs for pseudo label generation, the mIoU of our pseudo-masks can reach 71.9\%.

\begin{figure*}[t]
	\centering
	\begin{center}
		\includegraphics[width=\linewidth]{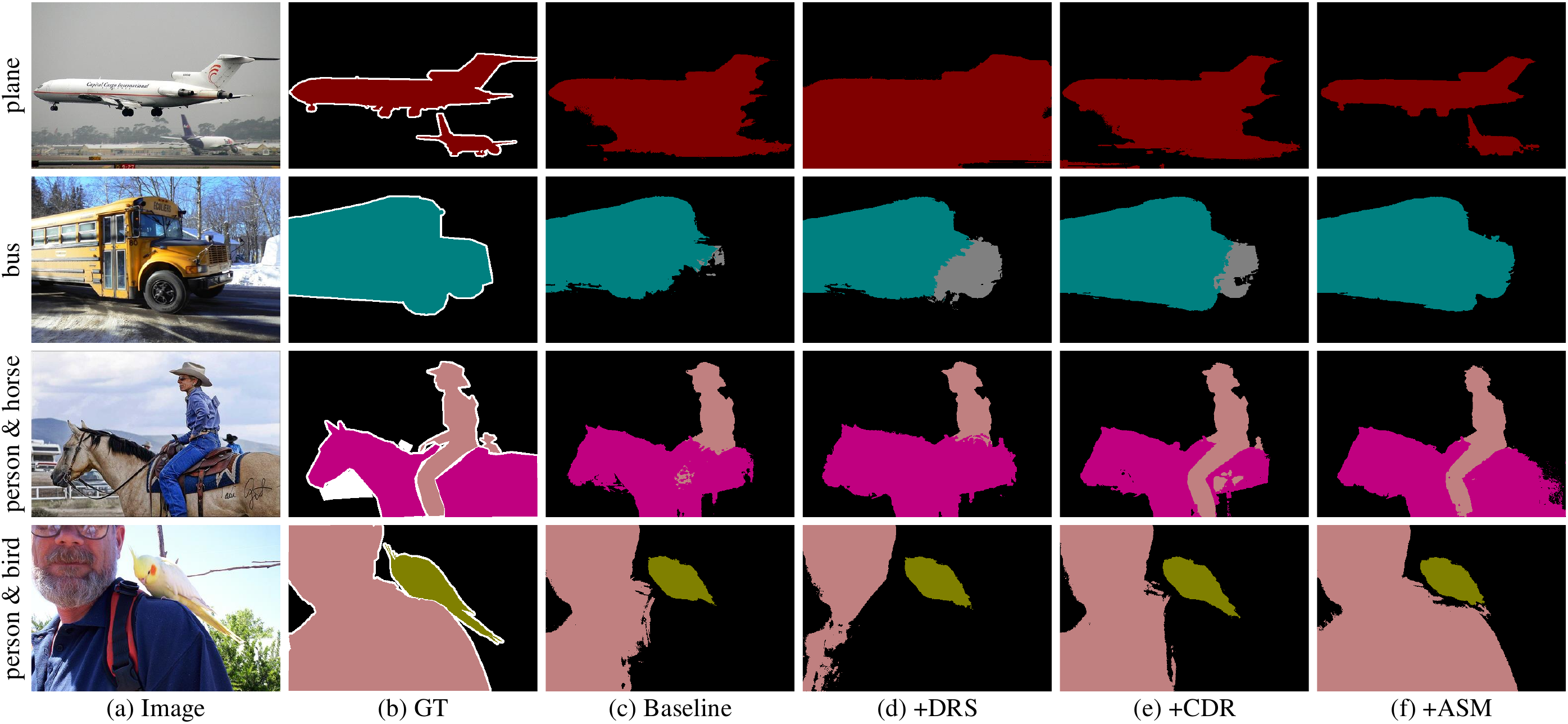}
	\end{center}
	\caption{Example prediction maps on the PASCAL VOC 2012 val set. For each (a) image, we show (b) the ground truth (GT), predictions of (c) baseline, (d) DRS \cite{kim2021discriminative}, (e) DRS + CDR, and (f) DRS + CDR + ASM.  Best viewed in color.}
	\label{fig_val}
\end{figure*}

\textbf{Accuracy of Segmentation Maps on PASCAL VOC 2012.} We present our segmentation results on PASCAL VOC 2012 for the backbone of VGG and ResNet in Table \ref{tab_vgg} and Table \ref{tab_resnet}, respectively. As can be seen, for the VGG backbone, our approach achieves better results than other state-of-the-art methods with only image-level labels. Specifically, our segmentation results can reach 67.4\% on the test set, which outperforms the recent work of ECS \cite{sun2021ecs} by 4\%. Our performance is also competitive to many approaches that rely on saliency maps, \eg, NSROM \cite{yao2021non} and EPS \cite{lee2021railroad}. For the ResNet backbone, we can get 72.7\% on the validation set and 72.8\% on the test set. Our proposed approach can outperform the recent work of ACR \cite{cheng2023out} and BECO \cite{rong2023boundary} by 0.6\% on the validation set and 0.9\% on the test set. Though approaches of W-OoD \cite{lee2022weakly} and CLIP-ES \cite{lin2023clip} leverage additional out-of-distribution images and language supervision to help distinguish the foreground from the background, our method can still outperform them by 1.4\%. Besides, our proposed SSC can also achieve competitive performance with recent SOTA approaches like PPC \cite{du2022weakly} and RCA \cite{zhou2022regional} that require additional saliency models. Some example prediction maps on the PASCAL VOC 2012 val set can be viewed in Fig.~\ref{fig_val}.

\begin{table}[t]

	\setlength{\tabcolsep}{3mm}
	\renewcommand\arraystretch{1.1}
	\centering
	\caption{Quantitative comparisons to previous state-of-the-art approaches on PASCAL VOC 2012 val and test sets with VGG backbone. Sup: Supervision, I: Image-level label, S: Saliency maps.}
	\begin{tabular}{{l}*{4}{c}}
		\hline
		Methods &   Sup & Backbone & Val & Test\\
		\hline
		DRS \cite{kim2021discriminative}$_{\text {AAAI21}}$&I+S&VGG16&63.6&64.4\\
		GSM \cite{Li2021GroupWiseSM}$_{\text {AAAI21}}$&I+S&VGG16&63.3&63.6\\
		NSROM \cite{yao2021non}$_{\text {CVPR21}}$&I+S&VGG16&65.5&65.3\\
		EPS \cite{lee2021railroad}$_{\text {CVPR21}}$&I+S&VGG16&67.0&67.3\\
		L2G \cite{jiang2022l2g}$_{\text {CVPR22}}$&I+S&VGG16&68.5&68.9\\
		
		\hline
		
		AffinityNet \cite{ahn2018learning}$_{\text{CVPR18}}$&I&VGG16&58.4 &60.5\\ 
		ICD \cite{fan2020learning}$_{\text{CVPR20}}$&I&VGG16&61.2&60.9\\
		BES \cite{chen2020weakly}$_{\text{ECCV20}}$&I&VGG16&60.1&61.1\\
		ECS \cite{sun2021ecs}$_{\text{ICCV21}}$&I&VGG16&62.1&63.4\\	
		
		\hline
		\textbf{SSC (Ours)} &I&VGG16&\textbf{66.8} &\textbf{67.4}\\	
		
		\hline	
	\end{tabular}
	
	\label{tab_vgg}	
\end{table}

\begin{table}[t]

 	\setlength{\tabcolsep}{3mm}
	\renewcommand\arraystretch{1.1}
	\centering
 	\caption{Quantitative comparisons to previous state-of-the-art approaches on PASCAL VOC 2012 val and test sets with ResNet backbone. Sup: Supervision, I: Image-level label, S: Saliency maps, O: Out-of-distribution data, L: Language supervision.}
	\begin{tabular}{{l}*{4}{c}}
		\hline
		Methods &   Sup & Backbone & Val & Test\\
		\hline
		DRS \cite{kim2021discriminative}$_{\text {AAAI21}}$&I+S&R101&71.2&71.4\\
		EPS \cite{lee2021railroad}$_{\text {CVPR21}}$&I+S&R101&71.0&71.8\\
		AuxSegNet \cite{xu2021leveraging}$_{\text{ICCV21}}$&I+S&WR38&69.0&68.6\\
		PPC \cite{du2022weakly}$_{\text {CVPR22}}$&I+S&R101&72.6&73.6\\
		RCA \cite{zhou2022regional}$_{\text {CVPR22}}$&I+S&R101&72.2&72.8\\
		L2G \cite{jiang2022l2g}$_{\text {CVPR22}}$&I+S&R101&72.1&71.7\\
		W-OoD \cite{lee2022weakly}$_{\text{CVPR22}}$& I+O &R101& 69.8 & 69.9 \\
    	CLIP-ES \cite{lin2023clip}$_{\text {CVPR23}}$&I+L&R101&71.1&71.4\\
    	LPCAM \cite{chen2023extracting}$_{\text {CVPR23}}$&I+S&R101&71.8&72.1\\		  
		
		\hline
		
		IRN \cite{ahn2019weakly}$_{\text{CVPR19}}$&I&R50&63.5&64.8\\
		ICD \cite{fan2020learning}$_{\text{CVPR20}}$&I&R101&64.1&64.3\\
		
		CONTA \cite{zhang2020causal}$_{\text{NIPS20}}$&I&R101&66.1&66.7\\
		AdvCAM \cite{lee2021anti}$_{\text{CVPR21}}$&I&R101&68.1&68.0\\
		CDA \cite{su2021context}$_{\text{ICCV21}}$&I&WR38&66.1&66.8\\
		ECS \cite{sun2021ecs}$_{\text{ICCV21}}$&I&WR38&66.6&67.6\\
		RIB \cite{lee2021reducing}$_{\text{NIPS21}}$&I&R101&68.3&68.6\\
		AMR \cite{qin2022activation}$_{\text{AAAI22}}$&I&R101&68.8&69.1\\
		MCTformer \cite{xu2022multi}$_{\text{CVPR22}}$&I&WR38&71.9&71.6\\
		AFA \cite{ru2022learning}$_{\text{CVPR22}}$&I&MiT-B1& 66.0 &66.3\\
		SIPE \cite{chen2022self}$_{\text{CVPR22}}$& I &R101& 68.8 & 69.7 \\
		ReCAM \cite{chen2022class}$_{\text{CVPR22}}$&I&R101& 68.5 &68.4\\
		PPC \cite{du2022weakly}$_{\text {CVPR22}}$&I&R101&67.7&67.4\\
		ViT-PCM \cite{rossetti2022max}$_{\text{ECCV22}}$&I&R101&70.3&70.9\\
		Spatial-BCE \cite{rossetti2022max}$_{\text{ECCV22}}$&I&WR38&70.0&71.3\\
		AEFT \cite{yoon2022adversarial}$_{\text{ECCV22}}$&I&WR38&70.9&71.7\\
		ESOL\cite{li2022expansion}$_{\text{NIPS22}}$&I&R101&69.9&69.3\\
		TOCO \cite{ru2023token}$_{\text {CVPR23}}$&I&VIT-B&69.8&70.5\\
		OCR \cite{cheng2023out}$_{\text {CVPR23}}$&I&WR38&72.7&72.0\\       	
        ACR \cite{cheng2023out}$_{\text {CVPR23}}$&I&WR38&71.9&71.9\\    
		BECO \cite{rong2023boundary}$_{\text {CVPR23}}$&I&R101&72.1&71.8\\
		D2CAM\cite{wang2023treating}$_{\text{ICCV23}}$&I&R101&71.2&70.7\\
        MARS \cite{jo2023mars}$_{\text {ICCV23}}$&I&R101&70.3&71.2\\

		\hline
		\textbf{SSC (Ours)} &I&R101&\textbf{72.7} &\textbf{72.8}\\	
		
		\hline	
	\end{tabular}  

	\label{tab_resnet}	
\end{table}

\begin{table}[t]	
	
	\setlength{\tabcolsep}{3mm}
	\renewcommand\arraystretch{1.1}
	\caption{Quantitative comparisons to previous state-of-the-art approaches on COCO val set with VGG backbone. Sup: Supervision, I: Image-level label, S: Saliency maps.}
	\centering
	\begin{tabular}{{l}*{4}{c}}
		\hline
		Methods & Sup & Backbone & Val \\
		\hline

		DSRG \cite{huang2018weakly}$_{\text {CVPR18}}$ &I+S&VGG16&26.0\\ 
  		IAL \cite{wang2020weakly}$_{\text {IJCV20}}$ &I+S &VGG16 &27.7\\
            GWSM \cite{Li2021GroupWiseSM}$_{\text {AAAI21}}$ &I+S&VGG16 &28.4\\
		EPS \cite{lee2021railroad}$_{\text {CVPR21}}$&I+S&VGG16  &35.7\\
            I2CRC \cite{chen2022saliency}$_{\text {TMM22}}$&I+S&VGG16&31.2\\
  		RCA \cite{zhou2022regional}$_{\text {CVPR22}}$&I+S&VGG16&36.8\\
            MDBA \cite{chen2023multi}$_{\text {TIP23}}$&I+S&VGG16&37.8\\
            \hline
  		BFBP \cite{saleh2016built}$_{\text {ECCV16}}$&I&VGG16&20.4\\ 
		SEC \cite{kolesnikov2016seed}$_{\text {ECCV16}}$&I&VGG16&22.4\\ 
            CONTA \cite{zhang2020causal}$_{\text {NIPS20}}$&I&VGG16 &23.7\\
            \hline
            \textbf{SSC (Ours) } &I&VGG16&\textbf{38.1} \\
		\hline
	\end{tabular}
	
	\label{tab_coco_vgg}	
\end{table}

\begin{table}[t]	
	
	\setlength{\tabcolsep}{3mm}
	\renewcommand\arraystretch{1.1}
	\caption{Quantitative comparisons to previous state-of-the-art approaches on COCO val set with ResNet backbone. Sup: Supervision, I: Image-level label, S: Saliency maps, L: Language supervision.}
	\centering
	\begin{tabular}{{l}*{4}{c}}
		\hline
		Methods & Sup & Backbone & Val \\
		\hline
            AuxSegNet \cite{xu2021leveraging}$_{\text {ICCV21}}$ &I+S&WR38 &33.9\\
            L2G \cite{jiang2022l2g}$_{\text {CVPR22}}$&I+S&R101&44.2\\   
            CLIP-ES \cite{lin2023clip}$_{\text {CVPR23}}$&I+L&R101&45.4\\
            LPCAM \cite{chen2023extracting}$_{\text {CVPR23}}$&I+S&R101&42.1\\	
            \hline
            IRN \cite{ahn2019weakly}$_{\text {CVPR19}}$ &I&R101 &41.4\\
            SEAM \cite{wang2020self}$_{\text {CVPR20}}$ &I&WR38 &31.9\\
		CONTA \cite{zhang2020causal}$_{\text {NIPS20}}$&I&R50 &33.4\\
  		PMM \cite{li2021pseudo}$_{\text {ICCV21}}$ &I &WR38 &36.7\\
		OC-CSE \cite{kweon2021unlocking}$_{\text {ICCV21}}$ &I &WR38 &36.4\\
		CDA \cite{su2021context}$_{\text{ICCV21}}$&I&WR38&33.2\\
       	RIB \cite{lee2021reducing}$_{\text{NIPS21}}$&I&R101&43.8\\
  		MCTformer \cite{xu2022multi}$_{\text{CVPR22}}$&I&WR38&42.0\\
      	SIPE \cite{chen2022self}$_{\text{CVPR22}}$& I &R101& 40.6 \\
      	ESOL\cite{li2022expansion}$_{\text{NIPS22}}$&I&R101&42.6\\
            TOCO \cite{ru2023token}$_{\text {CVPR23}}$&I&VIT-B&41.3\\
		OCR \cite{cheng2023out}$_{\text {CVPR23}}$&I&WR38&42.5\\    
      	ACR \cite{cheng2023out}$_{\text {CVPR23}}$&I&WR38&45.3\\  
		BECO \cite{rong2023boundary}$_{\text {CVPR23}}$&I&R101&45.1\\
		D2CAM\cite{wang2023treating}$_{\text{ICCV23}}$&I&R101&44.0\\
		\hline
            \textbf{SSC (Ours) } &I&R101&\textbf{47.0} \\
		\hline
	\end{tabular}  
	
	\label{tab_coco_resnet}	
\end{table}

\begin{figure}[t]
	\centering
	\begin{center}
		\includegraphics[width=\linewidth]{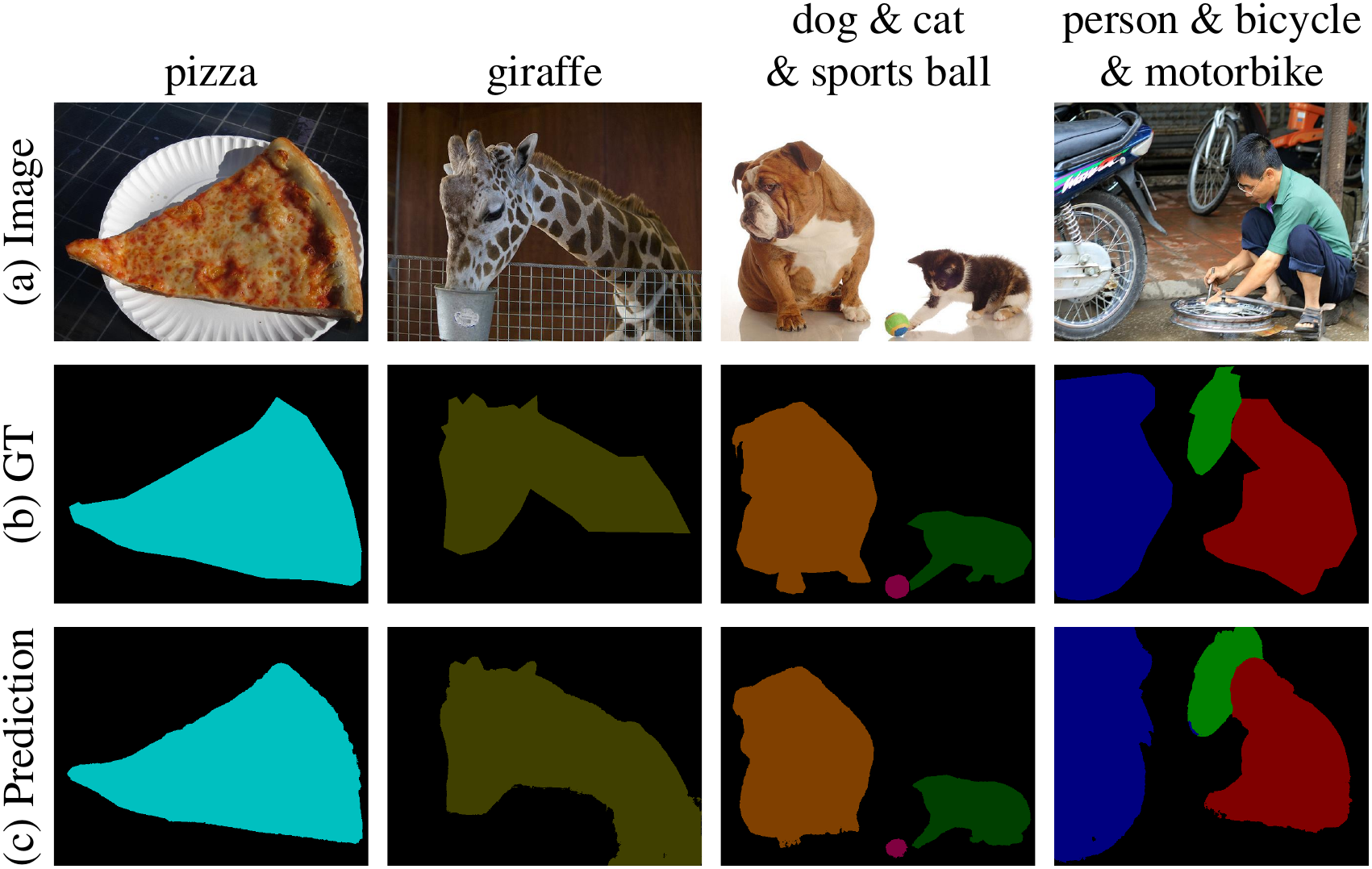}
	\end{center}
	\caption{Example prediction maps on the COCO val set. For each (a) image, we show (b) the ground truth (GT),  and (c) prediction. Best viewed in color.}
	\label{fig_val_coco}
\end{figure}

\begin{table}[t]
	
	\setlength{\tabcolsep}{4mm}
	\renewcommand\arraystretch{1.1}
	\centering
 	\caption{Element-wise component analysis. The accuracy (mIoU) of pseudo-masks on PASCAL VOC 2012 training set is reported. DRS \cite{kim2021discriminative}: Discriminative Region Suppression; CDR: CAM-Driven Reconstruction; ASM: Activation Self-Modulation; RAS: Reliable Activation Selection.}
	\begin{tabular}{{l}*{1}{c}}
		\hline
		Methods  & mIoU \\
		\hline
		baseline &48.3\\
		baseline + DRS  &50.2\\
		\hline
		\textcolor{gray}{baseline + DRS  + CDR early} &\textcolor{gray}{52.3}\\
		baseline + DRS  + CDR &54.5\\
		\textcolor{gray}{baseline + DRS  + CDR + ASM w/o RAS }&\textcolor{gray}{57.2}\\
		baseline + DRS  + CDR + ASM&\textbf{58.3}\\	
		\hline	
	\end{tabular}

	\label{tab_element}	
\end{table}

\begin{figure*}[t]
	\centering
	\begin{center}
		\includegraphics[width=\linewidth]{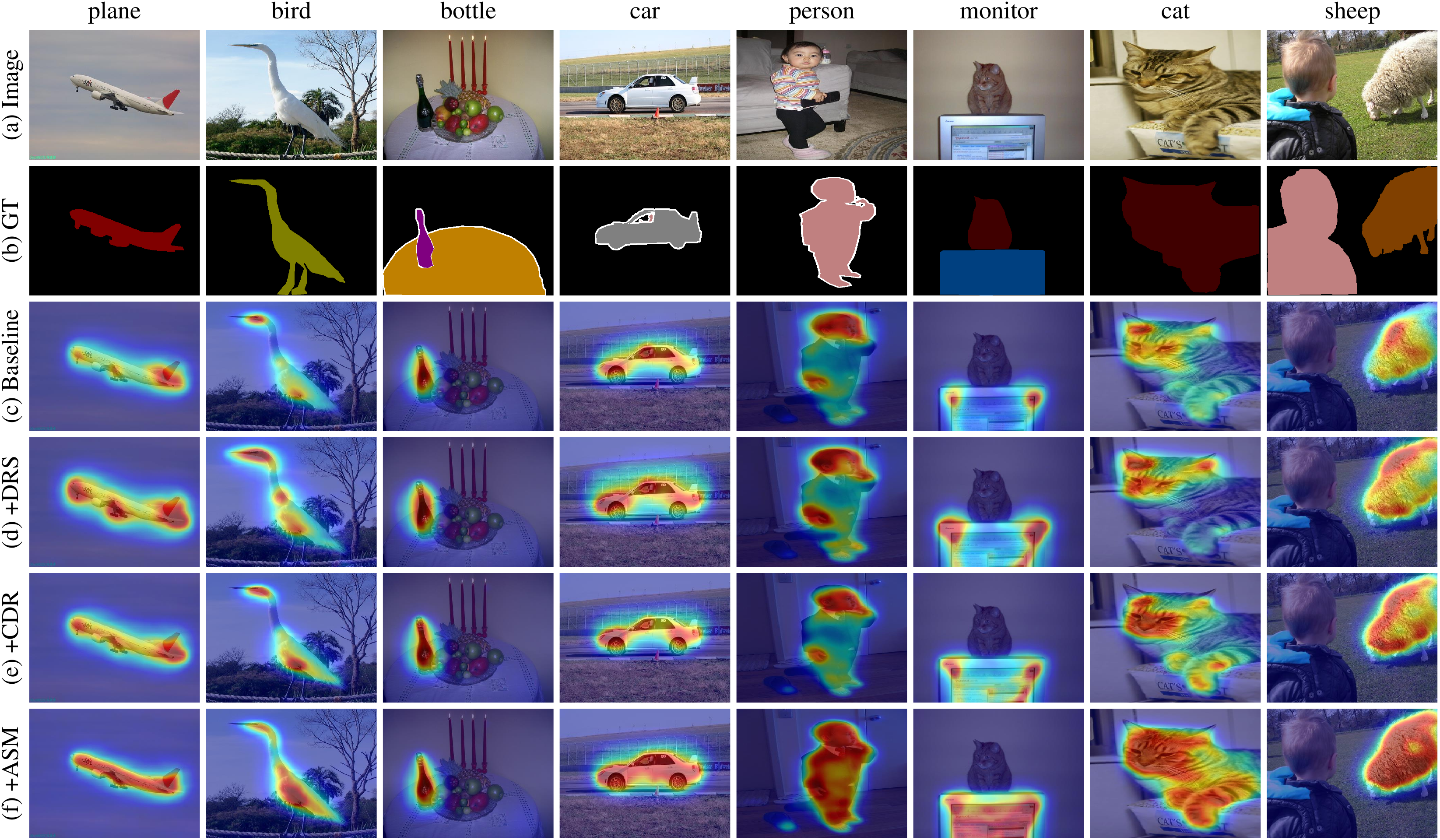}
	\end{center}
	\caption{Example localization maps on the PASCAL VOC 2012 training set. For each (a) image, we show (b) the ground truth (GT), localization maps produced by (c) baseline, (d) DRS \cite{kim2021discriminative}, (e) DRS + CDR, and (f) DRS + CDR + ASM.  Best viewed in color.}
	\label{fig_cam}
\end{figure*}

\begin{figure}[t]
	\centering
	\begin{center}
		\includegraphics[width=\linewidth]{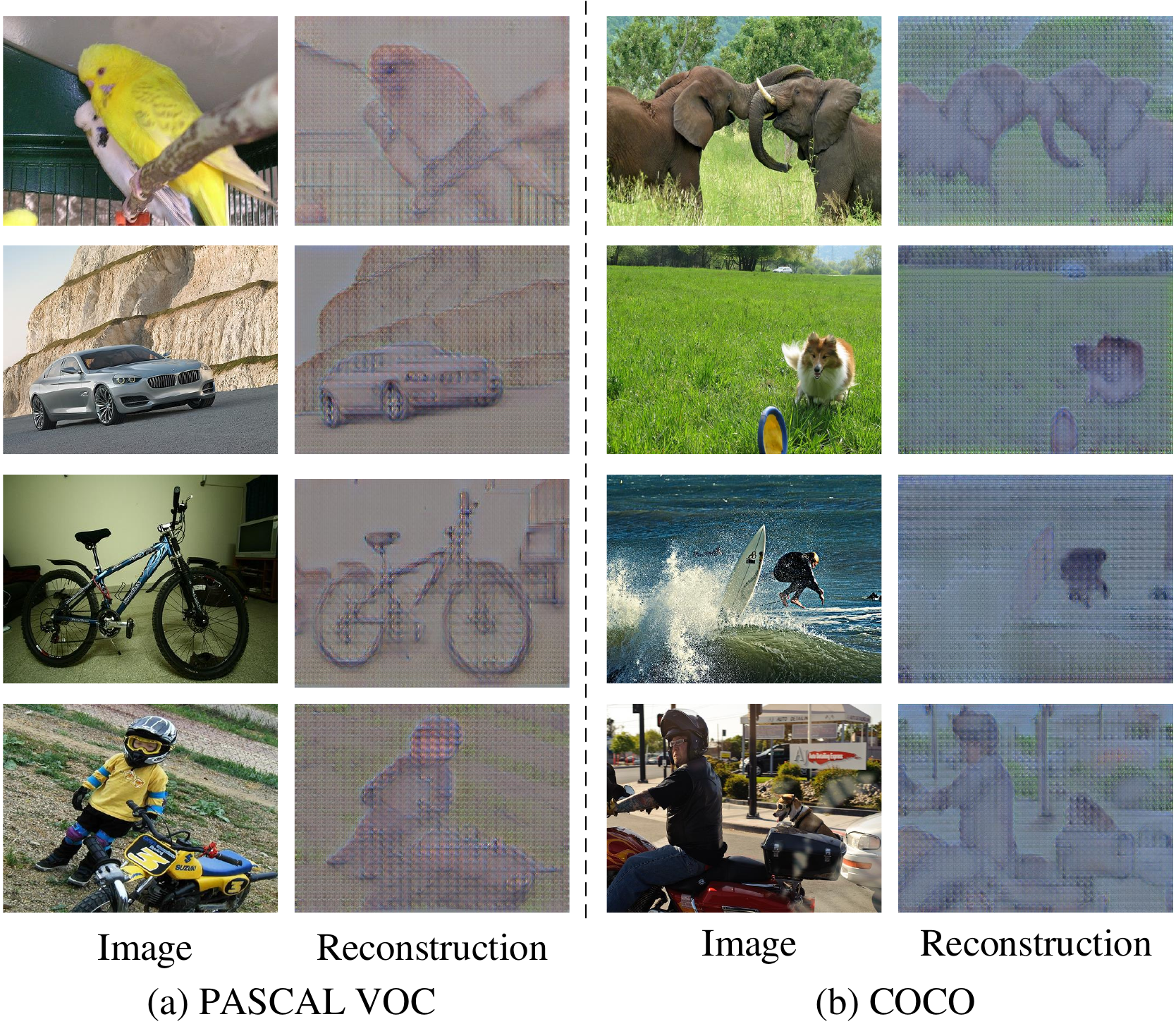}
	\end{center}
	\caption{Example image reconstruction results on the PASCAL VOC 2012 and COCO datasets. Best viewed in color.}
	\label{fig_rec}
\end{figure}

\textbf{Accuracy of Segmentation Maps on COCO.} For the more challenging COCO dataset, we provide performance comparisons with state-of-the-art WSSS methods for the backbone of VGG and ResNet in Table \ref{tab_coco_vgg} and Table \ref{tab_coco_resnet}, respectively. As shown in Table~\ref{tab_coco_vgg}, our proposed SSC with VGG backbone can achieve the performance of 38.1\% mIoU, much better than previous methods supervised with only image-level labels, \eg, 14.4\% mIoU higher than CONTA \cite{zhang2020causal}. Besides, our approach can also obtain better results compared to previous SOTA methods with additional saliency guidance (\eg, outperform RCA \cite{zhou2022regional} and MDBA \cite{chen2023multi} by 1.3\% and 0.3\% mIoU on the validation set, respectively ). Similarly, with the ResNet backbone, our proposed SSC reaches the best results of 47.0\% mIoU compared to previous SOTA WSSS methods. Specifically, our approach outperforms ACR \cite{cheng2023out} and BECO \cite{rong2023boundary} by 1.7\% and 1.9\% mIoU, respectively. Our SSC can also obtain 1.6\% higher mIoU than CLIP-ES \cite{lin2023clip} than relies on language-image pretraining models with super-large datasets. Some example prediction maps on the COCO val set can be viewed in Fig.~\ref{fig_val_coco}.

\subsection{Ablation Studies}
\textbf{Element-Wise Component Analysis.}
In this part, we demonstrate the contribution of each component proposed in our approach to improving the quality of pseudo-masks. The experimental results are given in Table \ref{tab_element}. While the accuracy of pseudo-labels from baseline is 48.3\%, DRS \cite{kim2021discriminative} can improve the result to 50.2\%. With our proposed CAM-driven reconstruction module, we can significantly improve the quality of pseudo-masks and the accuracy arrives at 54.5\%. We can notice that if we reconstruct the input image from early features, the accuracy will drop to 52.3\%. This highlights the importance of directly reconstructing the input image from its CAM features to preserve the spatial structure in attention maps. Some visualization of the image reconstruction results can be observed in Fig.~\ref{fig_rec}. As can be seen, the reconstructed results with the perceptual loss focus more on the content and spatial structure of images rather than low-level pixel information. With our activation self-modulation module, we further constrain the localization maps with detailed spatial structure information and improve the accuracy of pseudo-masks to 58.3\%. As can be seen, without our reliable activation selection strategy, the accuracy will decline to 57.2\%. This highlights the importance of maintaining the high activation of the most discriminative regions to alleviate the false modulation in the opposite direction.

Some example localization maps on the PASCAL VOC 2012 training set can be viewed in Fig.~\ref{fig_cam}. As can be seen from the first two columns, though DRS \cite{kim2021discriminative} can help enlarge the activated region, it will also inevitably result in over-activation where the expanded object activation intrudes into the background area. With our proposed CAM-driven reconstruction module, we can help constrain the activation within the object region (\eg, the plane in the first column and the beak of bird in the second column). Our proposed activation self-modulation module further constrains the network attention on target objects by enhancing regional consistency, which leads to more compact object activation. In addition, benefiting from our spatial structure constraints, our proposed CAM-driven reconstruction and activation self-modulation modules can also help activate more integral object regions to alleviate the remaining under-activation problem in difficult images, illustrated in the last column. Some example prediction maps on the PASCAL VOC 2012 val set can be viewed in Fig.~\ref{fig_val}.

\begin{table}[t]
	
	\setlength{\tabcolsep}{4mm}
	\renewcommand\arraystretch{1.1}
	\centering
	\caption{Comparison of perceptual loss with L1 and L2 losses for CAM-driven reconstruction (CDR). The accuracy (mIoU) of pseudo-masks on the PASCAL VOC 2012 training set is reported.}
	\begin{tabular}{{l}*{1}{c}}
		\hline
		Methods  & mIoU \\
		\hline
		w/o CDR & 50.2\\
		\hline
		w/ CDR + L1 loss&51.3\\
		w/ CDR + L2 loss&50.9\\
		w/ CDR + perceptual loss&\textbf{54.5}\\
		
		\hline	
	\end{tabular}
	
	\label{tab_comp_l1}	
\end{table}

\textbf{Discussion of Perceptual Loss.}
\label{sec:dis}
For driving the classification network to preserve the spatial structure of image content, we propose a CAM-driven reconstruction module to recover the input image. Then the perceptual loss is adopted to penalize the reconstructed image that semantically and spatially deviates from the original image. The motivation for selecting perceptual loss rather than L1 or L2 losses which are also widely used in self-supervised learning is twofold. First, L1 and L2 losses highlight the alignment of low-level information. Such strict per-pixel alignment will encourage the network to focus more on color and texture than image content and spatial structure. Second, image reconstruction with L1 or L2 loss requires the network to preserve more subtle information than perceptual loss, which might exceed the capacity of CAM features and thus deteriorate its localization ability. The qualitative comparison of perceptual loss with L1 and L2 losses for CAM-driven reconstruction is demonstrated in Table \ref{tab_comp_l1}. As can be seen, only about 1\% performance gain is achieved with L1 or L2 losses. In contrast, with the perceptual loss, our proposed CAM-driven reconstruction can improve the accuracy from 50.2\% to 54.5\%.

\begin{figure}[t]
	\centering
	\captionsetup[subfloat]{font=scriptsize}
	\subfloat[Weight of loss]{\includegraphics[width=0.24\textwidth]{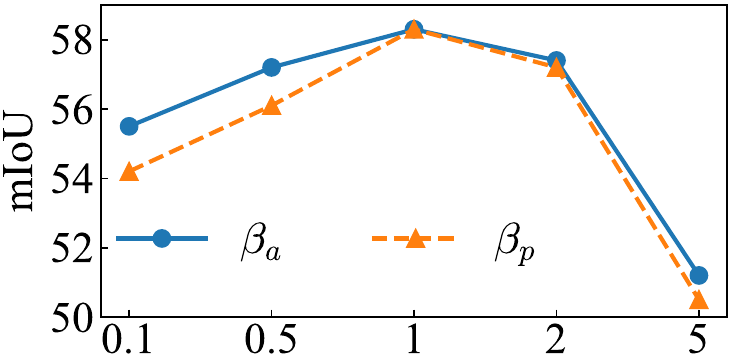}}
	\hfill 	
	\subfloat[Erosion kernel size]{\includegraphics[width=0.24\textwidth]{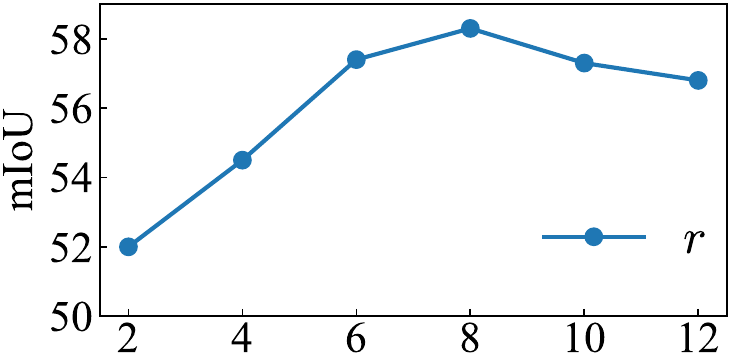}}
	\caption{The analysis of parameter sensitivity.  The accuracy (mIoU) of pseudo-masks on the PASCAL VOC 2012 training set is reported.}
	\label{fig_parameter}
\end{figure}


\textbf{Parameter Analysis.} 
For the CAM-driven reconstruction module, we conduct experiments to study the effect of perceptual loss. As shown in Fig.~\ref{fig_parameter} (a), we vary the weight $\beta _{p}$ over the range $\left \{0.1, 0.5, 1, 2, 5\right \}$. As we can see, we get better performance when the weight is between 1 and 2. A too-small $\beta _{p}$ may not improve the results very much, and a large $\beta _{p}$ deteriorates the performance seriously. We conjecture that increasing the importance of perceptual loss will make the network overemphasize reconstruction results and weaken the localization ability. Similarly, we vary $\beta _{a}$ with the same range to study the effect of the activation self-modulation module. Though more stable performance can be achieved while $\beta _{a}$ is between 0.1 and 2, the results drop significantly with a large weight as well. In our experiments, we empirically set $\beta _{p}$ = 1 and $\beta _{a}$ = 1.

For the reliable activation selection strategy, the object threshold $T_{obj}$ will affect the parameter of the erosion operation. Therefore, we first empirically fix $T_{obj}$ = 0.3 to locate the area with relatively higher attention values, which is a threshold adopted in OAA+ \cite{jiang2019integral} and NSROM \cite{yao2021non} as the foreground object threshold. Then we conduct experiments to study the effect of the erosion kernel size $r$. As shown in Fig.~\ref{fig_parameter} (b), we vary the kernel size $r$ over the range $\left \{2, 4, 6, 8, 10, 12\right \}$. As we can see, we get better performance when the kernel size is between 6 and 10. A too-large or small kernel size may not improve the results very much. We conjecture that a too-small kernel size keeps too much high activation in the CAM and mistakes the background as foreground, which will deteriorate the CAM modulation. Meanwhile, a too-large kernel size keeps less object activation, which dilutes the effect of reliable activation selection. In our experiments, we empirically set $T_{obj}$ = 0.3 and $r$ = 8.

\subsection{Limitation and Failure Cases.} 
\textbf{GPU Consumption.}
 As can be seen in Table \ref{tab_gpu}, compared to the parameter-free DRS \cite{kim2021discriminative},  the introduction of our CAM-driven reconstruction module will increase the GPU consumption from 3.2G to 10.6G. However, our full approach consumes only 14.2G GPU memory, which can be trained with a single 16G or 24G GPU.

\begin{figure}[t] 
 	\begin{center}
 		\includegraphics[width=\linewidth]{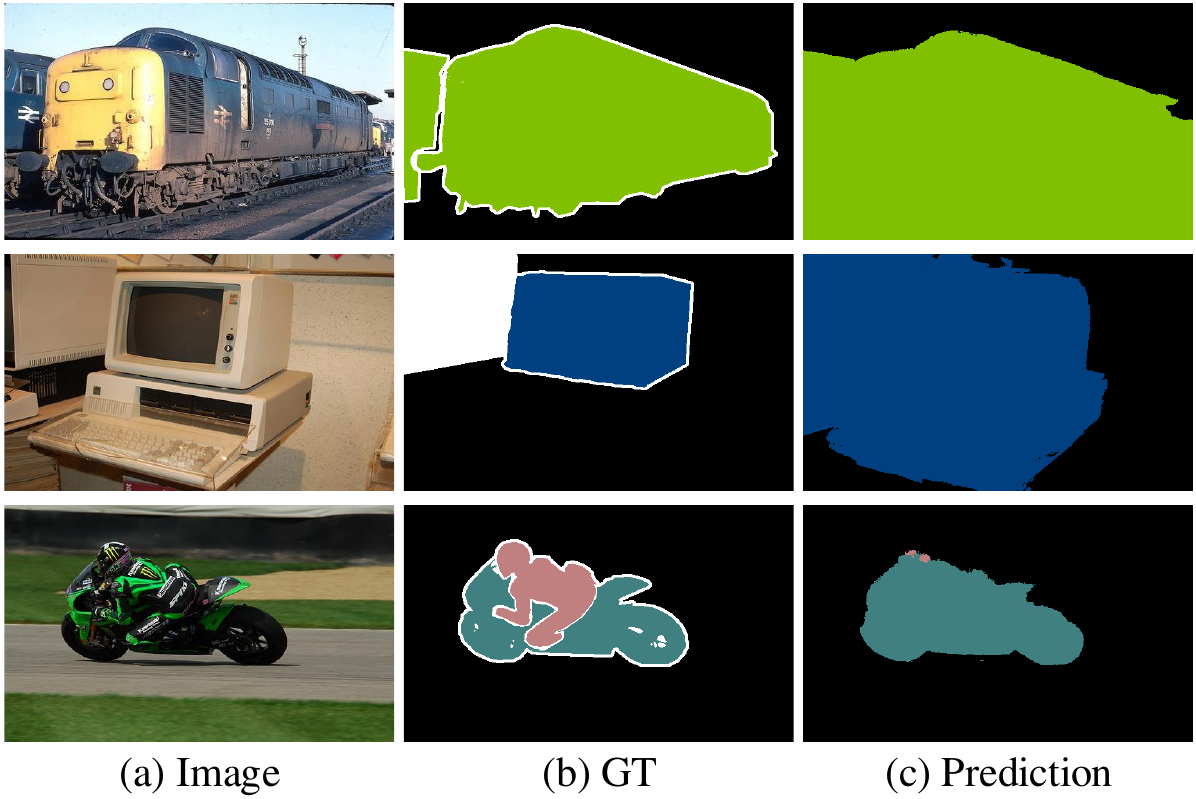}
 	\end{center}
 	\caption{Failure cases on PASCAL VOC 2012 val set. Best viewed in color.}
 	\label{fig_failure}
 \end{figure}

\textbf{Failure Cases.}
 Though our proposed SSC has achieved great success in the WSSS task, it still faces difficulties in some challenging scenes. Some typical failure cases are demonstrated in Fig.~\ref{fig_failure}. First, our approach may fail to tackle the co-occurrence problem. For example, in the 1st row of Fig.~\ref{fig_failure}, our model fails to distinguish the train from the railway background since these two categories often co-occur in images. Similarly, our method will treat the keyboard as the part of the monitor in the 2nd row. In addition, our model sometimes cannot recognize objects of different categories with similar appearances, \eg, it mistakes the person in the last row for the motorcycle. In the future, we would like to explore the solutions to the above challenges.
 
 \begin{table}[t] 
 	
 	\setlength{\tabcolsep}{6mm}
 	\renewcommand\arraystretch{1.1}
 	\centering
 	\caption{GPU consumption analysis. DRS \cite{kim2021discriminative}: Discriminative Region Suppression; CDR: CAM-Driven Reconstruction; ASM: Activation Self-Modulation.}
 	\begin{tabular}{{l}*{1}{c}}
 		\hline
 		Methods  & GPU \\
 		\hline
 		baseline &\multirow{2}*{3.2G}\\
 		baseline + DRS  &\\
 		\hline
 		baseline + DRS  + CDR &10.6G\\
 		baseline + DRS  + CDR + ASM&14.2G\\	
 		\hline	
 	\end{tabular}
 	
 	\label{tab_gpu}	
 \end{table}
 
\section{Conclusions}
\label{conclusion}

In this work, we proposed spatial structure constraints for weakly supervised semantic segmentation to alleviate the object over-activation problem during attention expansion. Specifically, we proposed a CAM-driven reconstruction module that directly reconstructs the input image from its CAM features. A perceptual loss was adopted to encourage the classification backbone to preserve the coarse spatial structure of the image content. Moreover, we proposed an activation self-modulation module to further refine CAMs with finer spatial structure details through enhancing regional consistency. Our proposed approach can help activate more integral target areas and constrain the activation within object regions. Extensive experiments on the PASCAL VOC 2012 and COCO datasets demonstrated the superiority of our proposed approach.

\bibliographystyle{IEEEtran}
\bibliography{egbib}

\end{document}